\documentclass{article}

\usepackage{url}

\usepackage[nonatbib,preprint]{neurips_2020}
\usepackage[utf8]{inputenc} 
\usepackage[T1]{fontenc}    
\usepackage{hyperref}       
\usepackage{booktabs}       
\usepackage{amsfonts}       
\usepackage{nicefrac}       
\usepackage{microtype}      
\usepackage{mathtools}
\usepackage{algorithmic}
\usepackage[linesnumbered,ruled,vlined]{algorithm2e}
\usepackage{caption}
\usepackage{graphicx}
\usepackage{amssymb}
\usepackage{array, caption, floatrow, tabularx, makecell, booktabs,amsthm,threeparttable}%
\usepackage{xspace}         
\usepackage{cleveref}
\theoremstyle{definition}
\newtheorem{definition}{Definition}[section]

\newtheorem{example}{Example}
\usepackage{appendix}

\newcommand{\eg}[0]{\emph{e.g.},\xspace} 
\newcommand{\ie}[0]{\emph{i.e.},\xspace}
\newcommand{\etal}[0]{\emph{et al.}\xspace}

\title{Discrete Few-Shot Learning for Pan Privacy}

\author{%
  Roei Gelbhart \\
  School of Computing and Information Systems\\
  University of Melbourne\\
  \texttt{gelbhartroei@gmail.com} \\
  \And
  Benjamin I. P. Rubinstein \\
  School of Computing and Information Systems \\
  University of Melbourne \\
  \texttt{benjamin.rubinstein@unimelb.edu.au} \\
}

\begin{document}
\bibliographystyle{ieeetr}
\maketitle

\begin{abstract}
	In this paper we present the first baseline results for the task of few-shot learning of discrete embedding vectors for image recognition. Few-shot learning is a highly researched task, commonly leveraged by recognition systems that are resource constrained to train on a small number of images per class. Few-shot systems typically store a continuous embedding vector of each class, posing a risk to privacy where system breaches or insider threats are a concern. Using discrete embedding vectors, we devise a simple cryptographic protocol, which uses one-way hash functions in order to build recognition systems that do not store their users' embedding vectors directly, thus providing the guarantee of computational pan privacy in a practical and wide-spread setting.
\end{abstract}

\section{Introduction}

In few-shot learning a classifier is trained to learn previously unseen classes given very few instances per class, motivated by humans who successfully few-shot learn a diverse range of tasks. For example, given a single image of a person, we are able to recognise other images of the same person; we can identify new animals given a single image of them; learn new letters in alphabets we've never seen before; we can easily identify fingerprints by matching patterns of skin lines to a reference fingerprint, when granted access to a reference database. Few-shot learning is of significant interest to industry. For example, security systems that rely on users' bio-metrics for recognition can only practically receive a few instances per person upon enrolment. In support of these applications, great strides have been made recently in the field of few-shot learning~\cite{hoffer2015deep,vinyals2016matching,PrototypicalCode, finn2017model, ravi2016optimization, sung2018learning, chen2019closer, tian2020rethinking}.
However in processing sensitive user data, \eg for user recognition or access control, data privacy must be addressed. For more information about the security of authentication systems (both password and bio-metric based), we refer the reader to \cite[Chapter 3]{van2020computer}.

We focus our attention on two questions:

\vspace{-0.5em}

\paragraph{Question 1 - Targeted Data Reconstruction.} Given access to the few-shot learning system, can an attacker recreate users' original data, \eg fingerprints, irises, etc.? We note multiple ways in which an attacker could achieve such access: by hacking into the system; via an offline system that stores data at endpoints; or a government warrant that compels the service provider to release confidential user data. The notion of pan privacy introduced by Dwork \etal~\cite{dwork2010pan}, refers to algorithms which can maintain privacy even if their inner state is visible to an attacker. Could neural network-based systems be made pan private? 
	This question can be motivated by extant attacks, for example the demonstration of Hitaj \etal~\cite{hitaj2017deep} in which training images of faces can be regenerated from trained models using GANs.

We take inspiration from modern authentication systems which do not store user passwords directly, but instead only a one-way hash~\cite{goldwasser2015lecture} once salted. This ensures that even an administrator of the system cannot gain access to user passwords.

\vspace{-0.5em}

\paragraph{Question 2 - Indiscriminate Model Inversion.} Consider a model that is robust to attacks of the type demonstrated in \cite{hitaj2017deep}: a user's training fingerprint cannot be recovered \emph{directly} from a learned model.  If the output of the model is not protected, however, then outputs of the model--- \eg a user's feature vector in the few-shot learning setting---are possibly susceptible to adversarial examples~\cite{biggio2013evasion} as produced by the FGS attack~\cite{goodfellow2014explaining} for instance. Even for cases in which the attacker can not modify directly the data, such as in airports where a security officer might be overseeing data entry, we see from the work of \cite{athalye2017synthesizing}, that it is possible to create 3D objects as adversarial examples. So a fake fingerprint, or even facial accessories, might be enough to fool the net.   
While different to sensitive original data, valid and sensitive inputs could still be created that result in target outputs. Can we protect the outputs of our models?

\textbf{Related Work.} We observe two kinds of related past work on few-shot learning: (i) Works that rely on learning a transformation from the data space to a Euclidean space, and subsequently classifying based on proximity; (ii) Approaches not based on Euclidean embeddings. For (ii), we highlight some  impressive results, such is the work of~\cite{finn2017model, finn2018probabilistic} on meta-learning, in which neural networks are trained on a variety of learning tasks, so as to adapt to new types of learning tasks with only a few SGD steps. Sung \etal~\cite{sung2018learning}, train a first model to produce an embedding vector which they concatenate to the embedding vectors of each candidate in the target class. The second model---a relation module---receives these pairs of embedding vectors as input, and predicts similarity. The approach of our work is to hide the embedding vectors using one-way hash functions, and thus render impossible, use of first model outputs as input to a second neural network. For a good survey over these methods, we refer the reader to~\cite{chen2019closer}.

Generally, it seems difficult to ensure theoretical privacy guarantees of highly accurate models which retrain over private data, as neural networks are effectively capable of remembering training data. We thus focus only on approaches of type (i), which train the model solely on public data in the training stage, and then once private users' data is received, the model is unchanged. Such work can be seen in \cite{hoffer2015deep, koch2015siamese, snell2017prototypical, ye2018deep}. The main idea of such approaches is to train the model to transform vectors from data space to feature space. The loss function (detailed in Section~\ref{sec_train_cnn}) will encourage the model to bring closer instances from same classes, and create a margin between instances from different classes. 

To protect data we design a privacy-preserving  algorithm (Section~\ref{sec_hashing}), which utilises one-way hash functions . One-way  hash functions require exact matches in the input of the function for any type of proximity in the output. This requires that the output of the model be discrete. Work on discrete neural networks has been previously explored in separate contexts~\cite{Liu_2016_CVPR, Lai_2015_CVPR, NIPS2017_7210}; to the best of our knowledge, ours is the first related to few-shot learning, in which the neural network has never seen a class it tries to hash. Lai \etal~\cite{Lai_2015_CVPR}  evaluate precision-recall curves for hamming radius of up to two. However, for few-shot learning, achieving such low radii is extremely difficult---as experimentally demonstrated in Section~\ref{sec_train_cnn}. Even if a model can achieve such low distances, in order to protect the embedding vectors using one-way hash functions, we must achieve perfect matching for the same classes. To address this problem we introduce a probabilistic algorithm we term \emph{Random Coordinate Projection Hashing (RCPH)}, that can utilise models with even large hamming distances.

\textbf{Our Contributions} are summarised as: i)
 We design a private few-shot recognition algorithm using one-way hash functions. We introduce the RCPH algorithm which enables the use of imperfect matching neural networks.
ii) We offer first experimental results for discrete few-shot learning. Finally we analyse the accuracy of RCPH coupled with the learned model.
   
\section{Privacy-Preserving Hashing Algorithm} \label{sec_hashing}

Denote by $f(x)$ a neural network which associates vectors from image space $\mathcal{X}$, into a discrete feature space $\mathcal{V}$. In our experiments, we choose $\mathcal{V} = \left \{0,1 \right \}^{1024}$. $f$ is trained to bring closer (in hamming distance) instances of the same class, while separating instances of different classes. In this way, proximity in the image of $f(\cdot)$ may be used to \emph{match} pairs of input instances. If the trained net is able to achieve perfect matches, \ie zero hamming distance with high probability only for members of the same classes, then the task of preserving privacy is straightforward as described in Section~\ref{subsec_Perfect_matching_nets}. However, this condition is extremely difficult to guarantee. Accordingly we present in Section~\ref{subsec_non_matching_nets} a probabilistic algorithm named \emph{Random Coordinate Projection Hashing (RCPH)}, which w.h.p effectively matches non-zero hamming-distanced vectors, provided there is a large margin between correct instance distance, and incorrect instance distances. 

\begin{definition}{\textbf{One-Way Function} (\cite[Definition 2.2]{goldwasser2015lecture}).}
A one-way function, $h(x)$, is a function for which, for all $x$, $h(x)$ can be computed in probabilistic polynomial time (PPT), but for every PPT algorithm, given $y=h(x)$, the probability of finding any source of $y$, $z$ for which $f(z)=y$ is negligible. Examples of one-way functions can be found in~\cite{goldwasser2015lecture}. 
\end{definition}

\subsection{Perfect Matching Nets} \label{subsec_Perfect_matching_nets}


Denote by $h(v)$ a one-way hash function, from $\mathcal{V}$ to a  hashing space $\mathcal{H}$. A simple privacy-preserving technique is the following: upon enrolment of a new user with input $x_i$ to the system (learning a new class $y_i$), the system saves $h(f(x_i))$, instead of $f(x_i)$. This provides a form of `computational privacy' provided that the original embedded $f(x_i)$ cannot be generated with computational efficiency. Such an approach would therefore provide pan privacy as its internal state would not reveal sensitive data. Upon testing a given instance $x'$, we search for an exact match in $\{h(f(x_1)),\ldots,h(f(x_{i_i}))\}$ for $h(f(x'))$, which likely exists only if the network model $f(\cdot)$ returns exact matches w.h.p. 


\textbf{Double hashing with ZKP verification.} Consider a setting where the system is deployed locally on many end-point machines. In such a case, for the sake of pan privacy, we should assume that $h(f(x_i))$ is publicly known. While an attacker may not be able to reconstruct $x_i$ or even $f(x_i)$ given $h(f(x_i))$, they might be able to break into another machine using $h(f(x_i))$, posing as someone else (the user who enrolled $x_i$ in the first place). For example, suppose that the protocol is to compute $f(x_i)$  on a local endpoint, and then $h(f(x_i))$ is sent to the server. If an intruder knows $h(f(x_i))$ (assuming it is public), they can directly send $h(f(x_i))$ to the system and login, without knowing $x_i$ or $f(x_i)$. To prevent this from happening, we modify the system, to save $h^{2}(f(x_i))$, as the user's ID for matching (which will be public), and $h(f(x_i))$ as a secret password. The password will not be saved anywhere, but only a zero-knowledge proof (ZKP) verifier will be stored, such as the one described in~\cite{bellare2002gq,kiefer2014zero}. Upon accessing the system, the users will identify themselves with $h^{2}(f(x_i))$, and authenticate by proving to the verifier that they know $h(f(x_i))$. ZKP protocols~\cite{goldwasser1996lecture,goldwasser1989knowledge,kiefer2014zero} can verify correctness of the password without storing or transmitting any information about the password itself.      

\subsection{Imperfect Matching Nets} \label{subsec_non_matching_nets}

Training perfect matching nets, that simultaneously achieve high accuracy, is a significant challenge that remains open. Thus, we take a different tact, motivated by the following example that demonstrates why perfect matching is not a necessity.

\begin{example}\label{eg:distance_vector}
Consider the matrix of a net's distances, as shown in Equation~\eqref{distance_vector}. We describe the net, applied to the Omniglot data set~\cite{lake2011one}, in Section~\ref{sec_train_cnn}. Each row of the matrix is the distance of a row-specific query instance, to each of the 10 classes' anchors---the hashed embeddings that would be enrolled by new users of the system, as described in the previous section. The bold distances represent the distance to the correct class for the query instance. We note that most of our experiments were performed with 20-way tasks, \ie with 20 options per query, but for compactness we illustrate just 10 here.  
\begin{equation}
M_{\mbox{dist}}=
\begin{pmatrix}
\textbf{7} & 35 & 50 & 28 & 34 & 45 & 28 & 31 & 49 & 37 \\
36 & 50 & 47 & \textbf{15} & 37 & 48 & 25 & 46 & 36 & 42 \\ 
36 & 25 & \textbf{30} & 38 & 26 & 39 & 21 & 30 & 32 & 26
\end{pmatrix}
\label{distance_vector}
\end{equation}
A system that stores embedding vectors without hashing, can calculate this distance matrix and return a nearest neighbour, which for the first two rows happens to be the correct class. However, a system that stores only hashes cannot. One-way hash functions send close vectors in the domain to arbitrary vectors in the co-domain, that do not preserve distance---an important distinction with \emph{locality-sensitive hashing}~\cite{indyk1998approximate} that serves approximate nearest neighbour search but that does not protect privacy. We can calculate these matrix rows when estimating the successes rate of our algorithm on a test set, but upon system deployment, this matrix is unknown and any effective system must make decisions without it.
\end{example}

\begin{algorithm}[t] 
	\caption{Random Coordinate Projection Hashing (RCPH)}
	\label{PM_algorithm}
	\SetKwInput{Parameters}{Parameters}
		\Parameters{ 

				$p$ the portion of coordinates to match,
				$m$ the maximum number of iterations.

	}

	\SetKwInput{preprocessed}{Preprocessed Data}
			\preprocessed{ 
				
				$n$ size of feature space,
				$k$ number of classes,
				$C=\{c_1,c_2,...,c_m\}$ set of random combinations,
				$H=\{h_1,h_2,...,h_m\}$ one-way hash functions,
				$A=\{A_1,A_2,...,A_k\}$ class anchors.  
	}

	\SetKwInput{Input}{Input}
\Input{ 

		$D_X \triangleq \{ h_1(f(x)|c_1),h_2(f(x)|c_2),..,h_i(f(x)|c_m) \}$ - hashes of combinations from $C$ of bits from the query's feature vector.
}

	\SetKwFunction{FSum}{RCPH$_{p,m}$}
	\SetKwProg{Fn}{Function}{:}{}
	\Fn{\FSum{$D_X$}}{
		
		\For{$h_i(x|c_i)$ in $D_X$ }{
			$index$ = Search($h_i(x|c_i)$ in $\{ h_i(a_1|c_i),h_i(a_2|c_i),..,h_i(a_k|c_i) \}$)
		
		\If{$index\neq None$}
		{
			\KwRet $index$
		}
	}
		\KwRet Abstain
	}
\end{algorithm}

\textbf{Random Coordinate Projection Hashing (RCPH)} is detailed in Algorithm~\ref{PM_algorithm} as the main algorithm of this paper. RCPH has two parameters $p,m$ that determine its accuracy, time and space complexities, and hashing space preservation. For this part of the paper we consider the following as inputs also: the neural network, the calculated anchors for all the classes (described in Section~\ref{sec_train_cnn}), the size of the feature space $n$, and the number of classes $k$. 

The algorithm iterates over hashes of partial combinations of bits from the embedding vector of a given input vector $x$, to the same combinations of bits from the anchors of the classes that were calculated at training time. Parameter $0<p\leq1$ is the portion of bits to match such that the number of bits that are randomly chosen at each iteration is $\lfloor p \cdot n \rfloor$. Here $m$ is the maximum number of iterations that the algorithm runs for. If after $m$ iterations no match is found, the algorithm chooses to abstain, thus limiting its time complexity. Each query can lead to one of three outcomes: correct match, wrong match, or abstention. In our analyses we will bound from below the average correct match rate, bound from above the average incorrect match rate, and bound from above the average time complexity. 

In the preprocessing stage $m$ random combinations of size $\lfloor p \cdot n \rfloor$ from $\{1,2,...,n\}$ are drawn, $C=\{c_1,c_2,...,c_m\}$, while $m$ different hash functions $H=\{h_1,h_2,...,h_m\}$ are generated. For each anchor of a class that was registered into the system, the algorithm stores $m$ hashes, choosing bits according to $C$. In iteration $i$, we compare a hash of bits $c_i$ from the query vector $f(x)$, denoted by $h_i(f(x)|c_i)$ to the same choice $c_i$ of bits from all the anchors, $\{ h_i(a_1|c_i),h_i(a_2|c_i),..,h_i(a_k|c_i) \}$, where $\{a_1,a_2,...,a_k\}$ denotes the anchor set. That is, we make comparisons of hashed random coordinate projections, as in
$h(v|c_i) = \left(\left(h(v)\right)_{c_{i1}},\ldots,\left(h(v)\right)_{c_{i\lfloor p\cdot n\rfloor}}\right)$. 

For each anchor $a_j$, we pre-compute during preprocessing the set $A_j = \{ h_1(a_j|c_1),h_2(a_j|c_2),..,h_i(a_j|c_m) \}$, caching them all in $A=\{A_1,A_2,...,A_k\}$ for use during algorithm execution.
We note that there could be two anchors that have the same partial hash for some $c_i \in C$. If that is the case, it means that either those two anchors are very close in embedding space (usually because of bad data, or a weak neural network), or that we have observed an unlikely collision event. In either case, the potential damage is quickly mitigated through randomising another coordinate projection. 
We emphasise that \textit{only hashes of partial embedding vectors are stored, thereby maintaining pan privacy.}

\subsection{RCPH Analysis} \label{sub_RCPH_Analysis}

Despite the fact that the distance vector (as illustrated by Example~\ref{eg:distance_vector}) of a query is unknown during activation, we can evaluate the accuracy of the algorithm using these vectors during test time. For example, for the given vector which we denote by $v$, and the correct label index, which we denote by $y$, we know that the probability of finding the correct match in an iteration, denoted by $\emph{E}_{c}$, is
\begin{equation}
\label{correct}
	\Pr( \emph{E}_{c}) =\frac{\binom{n-v(y)}{\left \lfloor pn \right \rfloor}}{\binom{n}{\left \lfloor pn \right \rfloor}}\approx (1-p)^{v(y)}\enspace.
\end{equation}
The approximation is valid when $n \gg v(y)$, but we do not use it in our code. The probability of having a wrong match in an iteration ($\emph{E}_{w}$) is bounded by the union bound,
\begin{equation}
\label{wrong}
\max_{i\neq y} \frac{\binom{n-v(i)}{\left \lfloor pn \right \rfloor}}{\binom{n}{\left \lfloor pn \right \rfloor}}
\leq
 \Pr( \emph{E}_{w}) \leq \sum_{i\neq y}  \frac{\binom{n-v(i)}{\left \lfloor pn \right \rfloor}}{\binom{n}{\left \lfloor pn \right \rfloor}}\enspace,
\end{equation}
and the probability of having no match in an iteration, denoted by $\emph{E}_{\emptyset}$ is similarly bounded as
\begin{equation}
\label{no_match}
1-\max_i \frac{\binom{n-v(i)}{\left \lfloor pn \right \rfloor}}{\binom{n}{\left \lfloor pn \right \rfloor}}
\geq
\Pr( \emph{E}_{\emptyset}) \geq 1 - \sum_{i}  \frac{\binom{n-v(i)}{\left \lfloor pn \right \rfloor}}{\binom{n}{\left \lfloor pn \right \rfloor}}\enspace.
\end{equation}

We noted before that during the pre-processing stage, we regenerate any random combination that created a collision between two anchors hashes, which should be in a good neural network, very rare. However, this will clearly complicate the analysis, and thus for simplicity, we assume that collisions can happen, and simply bound the probability of a single correct match ($\emph{E}_{sc}$) by 
\begin{equation}
\label{one_match}
 \Pr(\emph{E}_{sc}) \geq \Pr(\emph{E}_{c}) - \Pr( \emph{E}_{\emptyset})\enspace, 
\end{equation}
and conclude that 
\begin{align}
\Pr( \text{RCPH Correct}) = \sum_{i=0}^{m-1} \Pr(\emph{E}_{sc}) \cdot \Pr( \emph{E}_{\emptyset})^i \label{algo_success} \\
= \Pr( \emph{E}_{sc}) \frac{1-\Pr( \emph{E}_{\emptyset})^m}{1-\Pr(\emph{E}_{\emptyset})} \nonumber\enspace.    
\end{align}
Similarly, the wrong match probability can be bounded from above by
\begin{equation}
	\label{algo_fail}
	\Pr( \text{RCPH False})	= \Pr( \emph{E}_{w}) \frac{1-\Pr(\emph{E}_{\emptyset})^m}{1-\Pr(\emph{E}_{\emptyset})}\enspace.
\end{equation}

The time complexity of RCPH is bounded by $m$, but given the random nature of the algorithm, we can calculate its average case complexity. For each iteration, if we use a hash table to look for matches, the average time complexity is $O(1)$.  The probability of finding any match in one iteration is at least
$$
\lambda = \max_i \frac{\binom{n-v(i)}{\left \lfloor pn \right \rfloor}}{\binom{n}{\left \lfloor pn \right \rfloor}}\enspace,
$$
thus, by the expectation of a geometric random variable, the desired expectation is bounded by,
\begin{equation}
	\label{time_complexity}
\mathbb{E} T \leq \min (1/\lambda, m)\enspace.
\end{equation}
According to the law of total expectation we average over the expectations of all query points to obtain an estimate of the average-case time complexity. Using \Cref{correct,wrong,no_match,one_match,algo_success,algo_fail,time_complexity}, for each query we can bound from below the success rate shown in Equation~\eqref{algo_success}, and from above the fail rate and average time complexity, shown in Equations~\eqref{algo_fail}, \eqref{time_complexity}. Averaging over these values for an entire query set provides performance bounds of the system. 

\begin{table}[!htb]
\sffamily
  \floatsetup{floatrowsep=qquad, captionskip=4pt}
  \begin{floatrow}
    \ttabbox%
    {\begin{tabularx}{0.45\textwidth}{c *{2}{>{\centering\arraybackslash}X}}
      \toprule
      Accuracy & Fail Rate & Average Complexity \\
      \cmidrule(lr){3-3}\cmidrule(lr){1-1}\cmidrule(lr){2-2}
      0.9999 & 0.0000 & 130.66  \\
      \addlinespace
      0.2402 & 0.0002 & 10,000\\
    \addlinespace
      0.0000 & 0.0045 & 10,000\\
      \bottomrule
      \end{tabularx}}
    {\caption{p = 0.5, m = 10,000}\label{table1}}
    \hfill%
    \ttabbox%
        {\begin{tabularx}{0.45\textwidth}{c *{2}{>{\centering\arraybackslash}X}}
      \toprule
      Accuracy & Fail Rate & Average Complexity \\
      \cmidrule(lr){3-3}\cmidrule(lr){1-1}\cmidrule(lr){2-2}
      0.9995 & 0.0000 & 130.66  \\
      \addlinespace
      0.9345 & 0.0008 & 36,359\\
      \addlinespace
      0.0000 & 0.0445 & 100,000\\
      \bottomrule
      \end{tabularx}}
    {\caption{p = 0.5, m = 100,000}
      \label{table2}}
  \end{floatrow}
\end{table}%

In Tables~\ref{table1} and \ref{table2} we calculate the performance bounds for the three query points showed in the distance matrix of Example~\ref{eg:distance_vector}. The first point is the best one, as we can see that the correct label is within hamming distance 7. As demonstrated in our experiments, this is very much the common case. The second query is more borderline as we can see that the correct label is within distance 15, while the second best is relatively close in 25. We see that for $m=10000$ the accuracy is very poor, but if we are willing to pay with higher time complexity, the accuracy increases, and naturally so does the failure rate---more iterations mean higher chances for matching. We note that the numbers do not add up to one as they are bounds---one upper and one lower bound---while in addition the algorithm may abstain.

\subsection{Salting} \label{Salting}

A common practice in authentication systems is password salting. Given a user which has an ID and password, a random vector of bits is either added or concatenated to the password before it is hashed. Each user has their own salt vector, ensuring that users with identical passwords will have distinct hashed passwords after salting. If one uses large salt vectors, brute-force attacks need to be done per user, instead of over an entire password database. This mitigates amortising rainbow-table attacks.

In our case, salting appears infeasible. The embedding vector serves both as the identity and the password of the user, and thus we can not assign different salt values per user. However, if we are willing to relax that requirement, salting becomes a possibility.
For example, consider a system in which the user puts in their ID (their name for example), with their fingerprint serving as the password. Another option is using two neural networks in the process; the first will be used for ID recognition, and its data will be public (for example it will use the face of the user, or only a partial fingerprint), and the second will receive confidential data. For the first case we will use standard distance-based few-shot learning, and the data will not be protected, for the second case we use RCPH with salting.

When using salting, if we add a random binary vector of size $n$ to the feature vector, the distance between two different classes' feature vectors becomes on average $2^{\frac{n}{2}}$, and the probability of misclassification (fail rate) for reasonable $p,m$ values becomes negligible. 
So can we reduce $p$ (the required matching portion of the feature vector), to increase the accuracy? The answer depends on the quality of the data and the net. The salt is public knowledge, and thus an attacker that tries to break into a specific account, can add the salt as well. Reducing $p$ means that the attacker needs to guess a shorter part of the feature vector. If the attacker possesses another data point which is similar to the user's, they can brute-force over a small neighbourhood of that data point to hack in. Smaller $p$ means higher vulnerability to attacks. To reduce the vulnerability, we need to improve the net's separation abilities.

\section{Training Discrete Neural Networks for Few-Shot Learning} \label{sec_train_cnn}

\textbf{Data set.} We used the Omniglot data set \cite{lake2011one}, and the work on prototypical networks of \cite{snell2017prototypical} as a base line. We leveraged code from an implementation of \cite{snell2017prototypical} available at \cite{PrototypicalCode}. Omniglot contains 1623 characters from 50 different alphabets; each character having 20 different instances, each drawn by a different person. We use the same data splits as in \cite{vinyals2016matching,snell2017prototypical}, which consists of 90 degree rotations (considered to be different classes), which makes for a total of 6492 characters. The train, validation and test sets are of sizes 4112, 688 and 1692 respectively, and do not share characters from the same alphabet.  
 
 \begin{figure}[t]
	\includegraphics[width=\linewidth]{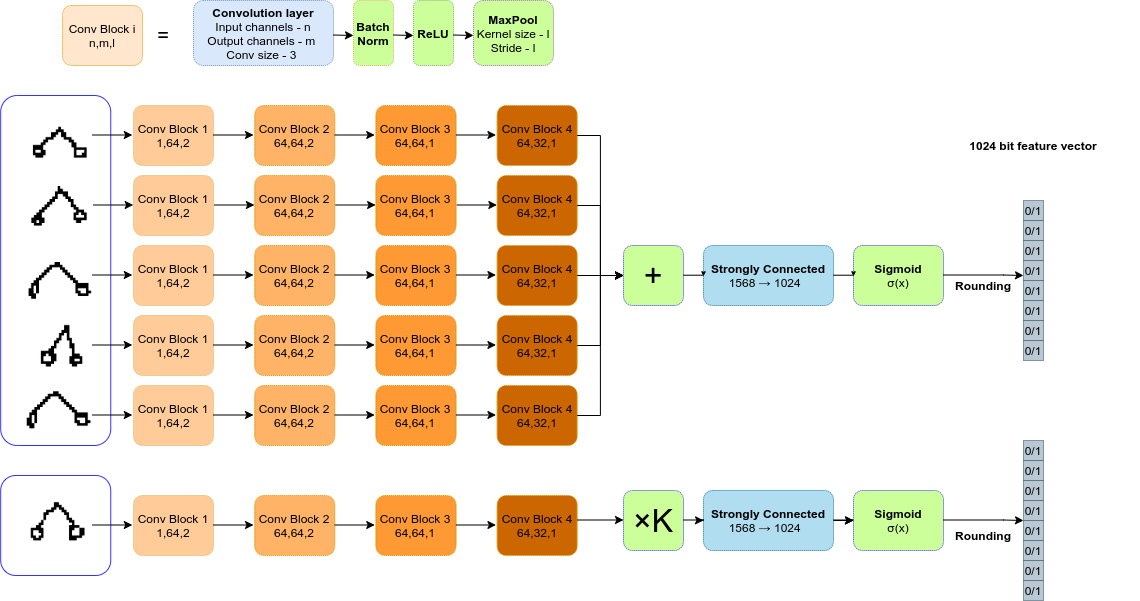}
	\centering
	\caption{Discrete Prototypical Network (DPN)}{Convolutional and fully connected blocks with same colours share weights. K in this example is 5, for one-shot learning is 1. }
	\label{model_figure}
\end{figure}

\textbf{Loss function.} We used softmax over the negated distance vector, to calculate the probability of each class. The loss is the negative log-likelihood. The number of classes we compare each query to is a hyper parameter. We chose 40 for the training stage, and for testing, we used 5 and 20 (corresponding to 5-way and 20-way few-shot learning).

\textbf{Model and output discretization.} Snell \etal~\cite{snell2017prototypical} introduced the idea of prototypical networks, which compute distances to prototype representations of each class, which are an average in feature space of examples in the support set of the class. The results from the original paper \cite{snell2017prototypical} are presented together with our results in Table~\ref{final_results}. Our model,  presented in Figure \ref{model_figure}, is very similar to the model in \cite{snell2017prototypical}, but with one major difference. Our goal is to encourage the net to output vectors which are close to binary vectors even before the final rounding, so as to ensure that the rounding of the vector will not significantly damage accuracy. In \cite{snell2017prototypical}, the main strategy for few-shot learning (demonstrated with 5-shot learning), was to average the 5 anchor instances of the class to a single anchor in feature space. If we try to copy their idea, and the net was able to output close to discrete values for the queries, the average of the anchors will likely be far from discrete: it will be discrete only if they are all identical. To solve this problem, we introduce another strongly-connected layer, that receives as input the sum of the anchors, which is equivalent to their average (up to a factor). In order to share the parameters of the anchors' net and query's net, we perform a small modification for queries, where we multiply the output of the final convolutional layer, by the number of anchors per class, as seen in Figure~\ref{model_figure}. 

Using strongly-connected layers, and a final sigmoid activation function, we are able to architecturally limit the values of the net output to be in $(0,1)^{1024}$. We refer to our architecture as \emph{Discrete Prototypical Networks (DPN)}. During training, rounding will eliminate the derivative, thus we must train without it. It might be beneficial to regularise the weights of the net such that the output from the sigmoid layer will be close to either 0 or 1, so that the difference between the artificial rounding, and the actual net's output will be as small as possible. To test this hypothesis, we added a regularization factor to the loss function, very similar to the one introduced in \cite{Liu_2016_CVPR}, $\lambda \left \| f(x) - \lfloor f(x)\rceil \right \|_2^2$.

We trained the model with different $\lambda$ values, and as result, the average of the distance of the net's output to the rounded output, which we refer to as ``the discretization gap", reduced from \textbf{$0.09$} without regularization, to \textbf{$0.06$} with the largest $\lambda$. However, the accuracy of the model decreased as $\lambda$ grows larger (graphs in the Appendix). Interestingly, the value of the discretization gap even without regularization seems rather small. We examined the gap as function of training time, and notice that even without regularization it decreases from \textbf{$0.21$} at start, to \textbf{$0.09$} at finish. It seems that the vanishing gradient phenomenon, which is very common for sigmoid activation functions, actually serves our purpose well. When the output of the net is close to 0/1 the gradient of the sigmoid is small, and thus small changes are made to the net.   

\begin{figure}[]
   \begin{floatrow}
     \ffigbox{\includegraphics[width=0.5\textwidth]{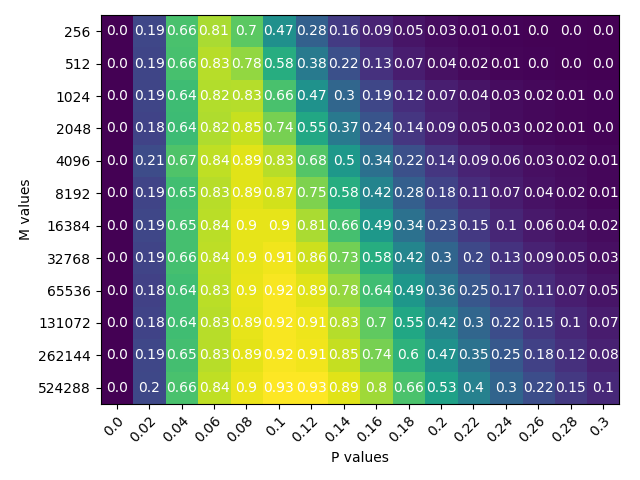}}{\caption{Accuracy lower bound.}\label{Accuracy}}
     \ffigbox{\includegraphics[width=0.5\textwidth]{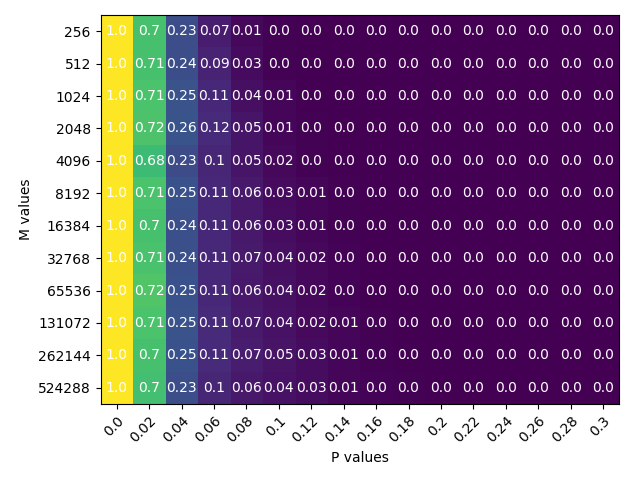}}{\caption{Failure rate upper bound.}\label{fail_rate}}
   \end{floatrow}
\end{figure}

\textbf{Experiments.} 
We present our final results in Table~\ref{final_results}. When dealing with security validation systems, $m$ of magnitude of $10^6$ is very reasonable, as the delay for the user is only upon entrance to the system, and with today's computing power, it is still well beneath a second. The average complexity, is usually significantly lower. In Figures~\ref{Accuracy} and \ref{fail_rate} we can see how the accuracy and failure rate behave as a function of $p,m$ for our net. We emphasise that using different training methods or architectures, which regulate the average best hamming distance, can significantly change the location of the best $p$. In this work we focused purely on maximising the accuracy of the net before discretizing. This is not necessarily the best approach, especially if one desires higher $p$ values (for better security).    

\begin{threeparttable}[t]
    {
    \begin{tabularx}{\textwidth}{l c *{8}{>{\centering\arraybackslash}X}}
          \toprule
      \toprule
        &\multicolumn{4}{c}{1-shot, 5-way} & \multicolumn{4}{c}{5-shot, 5-way} \\
              \cmidrule(lr){2-5}\cmidrule(lr){6-9}
       & Acc. & Fail Rate & Comp-lexity & Best P & Acc. & Fail Rate & Comp-lexity & Best P  \\
      \cmidrule(lr){2-5}\cmidrule(lr){6-9}
      \textbf{Prototypical Networks\tnote{*}\cite{snell2017prototypical}}  & 98.8\% & - & - & - & 99.7\% & - & - &-\\
      \addlinespace
      \textbf{DPN\tnote{*}} & 97.8\% & - & - & - & 99.4\% & - & - &-\\
      \addlinespace
      \textbf{DPN+RCPH, $\mathbf{m=10^3}$\tnote{**}} & 86.1\% & 6.5\% & 174& 0.07 & 94.2\% & 2.6\% & 82 &0.07 \\
      \textbf{DPN+RCPH, $\mathbf{m=10^4}$\tnote{**}} & 89.6\% & 5.2\% & 1182 & 0.09 & 95.9\% & 1.8\% & 547 & 0.09\\
      \textbf{DPN+RCPH, $\mathbf{m=10^5}$\tnote{**}} & 93.9\% & 3.7\% & 3961 & 0.10 & 97.5\% & 1.3\% & 1924 & 0.10\\
      \textbf{DPN+RCPH, $\mathbf{m=10^6}$\tnote{**}} & 94.5\% & 3.0\% & 42616& 0.12& 97.6\% & 1.3\% & 17152 &0.12 \\
      \end{tabularx}

    \begin{tabularx}{\textwidth}{l c *{8}{>{\centering\arraybackslash}X}}
      \toprule
        &\multicolumn{4}{c}{1-shot, 20-way} & \multicolumn{4}{c}{5-shot, 20-way} \\
        \cmidrule(lr){2-5}\cmidrule(lr){6-9}
       & Acc. & Fail Rate & Comp-lexity & Best P & Acc. & Fail Rate & Comp-lexity & Best P  \\
      \cmidrule(lr){2-5}\cmidrule(lr){6-9}
      \textbf{Prototypical Networks\tnote{*}\cite{snell2017prototypical}}  & 96.0\% & - & - & - & 98.9\%  & - & - &-\\
      \addlinespace
      \textbf{DPN\tnote{*}} & 93.6\% & - & - & - & 98.0\% & - & - &-\\
      \addlinespace
      \textbf{DPN+RCPH, $\mathbf{m=10^3}$\tnote{**}} & 68.9\% & 18.4\% & 270& 0.08 & 84.0\% & 8.6\% & 148 &0.08 \\
      \textbf{DPN+RCPH, $\mathbf{m=10^4}$\tnote{**}} & 77.0\% & 13.9\% & 1857 & 0.10 & 89.0\% & 5.4\% & 885 & 0.10\\
      \textbf{DPN+RCPH, $\mathbf{m=10^5}$\tnote{**}} & 81.7\% & 11.0\% & 13704 & 0.12 & 92.5\% & 4.1\% & 5832 & 0.12\\
      \textbf{DPN+RCPH, $\mathbf{m=10^6}$\tnote{**}} & 84.1\% & 9.6\% & 109302& 0.14& 93.4\% & 4.0\% & 27666 &0.13 \\
      \bottomrule
    \bottomrule
      \end{tabularx}
      }
      \caption{Results.}
    \label{final_results}
    \begin{tablenotes}
     \item[*] Without privacy - using nearest neighbour.
     \item[**] Computationally secure system. Given m, the ideal p is calculated over the validation set to maximise the accuracy, and then plugged in to the test set analysis.
   \end{tablenotes}
   \end{threeparttable}%

\section{Future Work}
As always for work that present baseline results for a new setting, our paper opens up a range of new directions for future research. 

\textbf{Better Accuracy.}
From Figures~\ref{Accuracy} and~\ref{fail_rate} we see the trade-off between higher accuracy and lower failure rate. Reducing $p$ lowers the probability of falsely entering the wrong person into the system (and as consequence revealing part of their feature vector). At the same time, the accuracy rises as well (it stops rising due in large part to the fact that we have a lower bound on accuracy, not an exact value). If we want to improve the trade-off, we need to improve the net---lower hamming distance for the correct class, and larger distance to the second best.

\textbf{Attacks and Defences.}
A brute-force attack against our model will require time complexity of $2^{n*p}$, which for our case is roughly $2^{1000*0.1} \approx 10^{30}$. However, an attacker that has access to data points from the same distribution, can investigate the outputs of the net to find correlation between bits, and thus reduce the brute-force search space. We call these ``data dependent Attacks", and they pose a serious threat to the security of our algorithm. The attacker can alternatively start from a specific data point, and search for close neighbours in hamming space. Given enough data points, the actual space that needs to be covered is a lot smaller than $10^{30}$, and dependent on the proximity in hamming space of the attacker queried data to the targeted data point. 

When dealing with one-way hash functions, the size of the input space determines the time complexity of brute force attacks. For our case, it is a subgroup of $\mathcal{V} = \left \{0,1 \right \}^{1024}$, but it is still not clear which points in $\mathcal{V}$ are reachable by the net from the data distribution, and what is the final probability distribution of the net's output over the data. If for example, the net utilises only the first half of the bits,  while the second half remains zero, or equal to the first half, then the output space is a lot smaller. How can we measure the randomness of the net's output, and how can we encourage the net to reduce correlation between output bits? We leave these questions for future work.

We note that the randomness of the net's output is limited not only by the net ability to discover features in the data, but also by the data itself; the dimension of the input data, and the number of actual features that are possible to detect.

\section*{Broader Impact}

Authentication systems based on computer vision and deep learning are prevalent. It is commonplace to use fingerprint or facial recognition to unlock our smartphones, while airport immigration is increasingly relying on  bio-metric data to reduce waiting times in passport control. Despite progress in safe storage of user passwords, AI-based systems introduce new vulnerabilities. First, the data itself might be valuable for some user. Second, even the mere output of the net (without data reconstruction) can be used in an adversarial attack to hack into users' accounts. In our work we present a first approach for a pan-private deep learning recognition system. We expect that research in this new direction will promote improved user privacy---a fundamental human right that should be afforded to citizens even when an authentication system is required for national security---and improved security---through improved safeguarding of credentials. We highlight data dependent attacks as a potential vulnerability and call for researchers to explore extensions that mitigate any risk from such attacks.



%
%

\bibliography{citations}

\newpage

{\huge\bfseries Appendix \par}

\appendix

\section{Further Results on Discretization}

\begin{figure}[h]
\label{fig:supp}
   \begin{floatrow}
     \ffigbox{\includegraphics[width=0.5\textwidth]{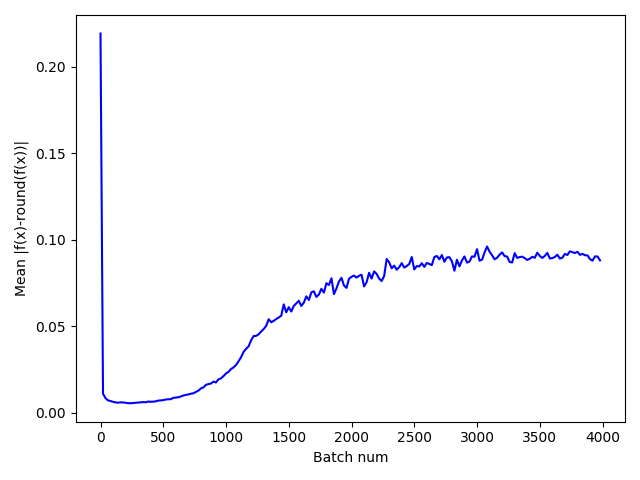}}{\caption{Natural discretization.}\label{case1}}
     
     \ffigbox{\includegraphics[width=0.5\textwidth]{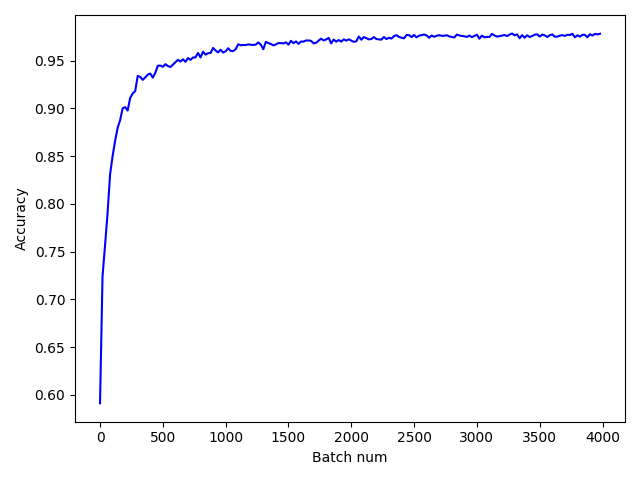}}{\caption{Accuracy as a function of time. }\label{case2}}
     
   \end{floatrow}

\label{fig:supp}
     \ffigbox{\includegraphics[width=0.7\textwidth]{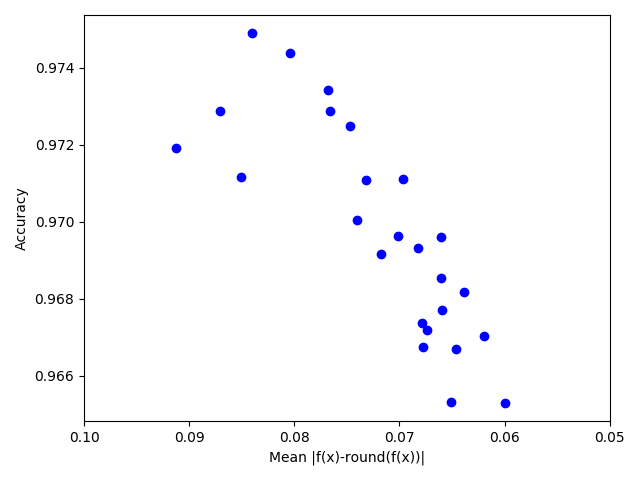}}{\caption{Accuracy as a function of regularization.}\label{acc_reg}}
\end{figure}

In this appendix we explore the effect of training our model with varying $\lambda$ values. The results, which can be seen in Figure~\ref{acc_reg}, show little influence over the accuracy, and in fact, the more we regularise, the worse the accuracy becomes. Interestingly, we can see from Figure~\ref{case1}, that even in a model without regularisation, that the average distance to the rounded output (the discretization gap), decreases with training in the first $2.5$ epochs (each epoch is $100$ batches, and it reduces to a minimum of $0.0053$ after $243$ batches). It appears that the vanishing gradient problem, which is very common for sigmoid activation functions, actually serves our purpose well. When the output of the net is close to 0/1 the gradient of the sigmoid is small, and thus only small changes are made to the net.   

An interesting phenomenon is the slow increase of the discretization gap after $2.5$ epochs. This increase lasts for approximately $30$ epochs, while the accuracy has already reached $0.91$. The net, which is obviously oblivious to the discretization gap, begins to utilise values in $(0,1)$ which are further from the edges.

\end{document}